\documentclass{article}

\usepackage[final]{nips_2016}

\usepackage{url}            
\usepackage{booktabs}       

\usepackage{amssymb}
\usepackage{amsmath}
\usepackage{mathtools}
\usepackage{bigdelim}
\usepackage{subcaption}
\usepackage{pgfplots}

\usepackage{times}
\usepackage{helvet}
\usepackage{courier}

\usepackage{bbm}
\usepackage{mathtools}
\usepackage{centernot}
\usepackage{enumerate}

\usepackage{pgfplots}
\usepackage[caption=false,font=footnotesize]{subfig}
\usepackage{undertilde}
\usepackage{verbatim}
\usetikzlibrary{arrows,calc}
\usepackage{relsize}

\usepackage{algorithm}
\usepackage{algpseudocode}

\usepackage{pst-node}
\usepackage{auto-pst-pdf}

\title{A Semi-Markov Switching Linear Gaussian Model for Censored Physiological Data}

\author{
Ahmed M. Alaa$^{\dagger}$, Jinsung Yoon$^{\dagger}$, Scott Hu$^{*}$ and Mihaela van der Schaar$^{\dagger}$ \\
$^{\dagger}$Electrical Engineering Department, $^{*}$David Geffen School of Medicine\\
University of California, Los Angeles\\
\texttt{ahmedmalaa@ucla.edu} \\
}

\begin{document}

\maketitle

\begin{abstract}
Critically ill patients in regular wards are vulnerable to unanticipated clinical deterioration which requires timely transfer to the intensive care unit (ICU). To allow for risk scoring and patient monitoring in such a setting, we develop a novel Semi-Markov Switching Linear Gaussian Model (SSLGM) for the inpatients' physiology. The model captures the patients' latent clinical states and their corresponding observable lab tests and vital signs. We present an efficient unsupervised learning algorithm that capitalizes on the informatively censored data in the electronic health records (EHR) to learn the parameters of the SSLGM; the learned model is then used to assess the new inpatients' risk for clinical deterioration in an online fashion, allowing for timely ICU admission. Experiments conducted on a heterogeneous cohort of 6,094 patients admitted to a large academic medical center show that the proposed model significantly outperforms the currently deployed risk scores such as Rothman index, MEWS, SOFA and APACHE.   
\end{abstract}
 
\section{The SSLGM Model}	
We focus on patients who are hospitalized and monitored in a regular ward in anticipation of potential clinical deterioration that may require an ICU admission. Patients are monitored via a set of $M$ physiological streams (i.e. vital signs and lab tests) which manifest their latent clinical states. The physiological measurements of every patient are gathered in discrete time steps $t \in \mathbb{N}$ (e.g. one measurement every 4 hours), and clinicians assess the patient's state --in real-time-- accordingly. At every time step $t$, a patient resides in 1 out of $N$ possible clinical states; each state reflects a certain level of severity of the patient's clinical condition. State indexes reflect the levels of clinical severity, i.e. state 1 is a {\it clinical stability state} in which a patient can be safely discharged from the ward, whereas state $N$ is a {\it clinical deterioration state} that requires transferring the patient urgently to the ICU. The clinical state-space is defined as $\mathcal{X} = \{1,.\,.\,.,N\}$; states are hidden, but manifest themselves through the physiological measurements, which are modeled as follows [1]
\begin{align}
Z_{t} &= A_{X_t} Z_{t-1} + B_{X_t} e_{t}, \nonumber \\
Y_{t} &= C_{X_t} Z_{t} + D_{X_t} w_{t}, \nonumber
\label{eq1}
\end{align}
where $Z_t \in \mathbb{R}^{M_z}$ is an $M_z$-dimensional {\it latent factor}, $Z_1|X_1 = x \sim \mathcal{N}(0, \Sigma_{x})$, $Y_{t} \in \mathbb{R}^{M}$ is a vector comprising the physiological data, $X_t \in \mathcal{X}$ is the patient's latent state, and for every $x \in \mathcal{X}$, $\{A_{x} \in \mathbb{R}^{M_z \times M_z}, C_{x} \in \mathbb{R}^{M \times M}\}$ are the (stable) matrices describing the linear dynamics, $e_{t}, w_{t} \sim \mathcal{N}(0, I)$ and $B_{x} = \mbox{diag}(\sigma^{2}_{b1},.\,.\,.,\sigma^{2}_{bM_z}), d_{x} = \mbox{diag}(\sigma^{2}_{d1},.\,.\,.,\sigma^{2}_{dM})$.   	

The clinical state sequence $\{X_t\}$ follows a semi-Markovian model; semi-Markovianity eliminates the unrealistic assumption of memoryless state transitions adopted by ordinary Markov chains. We adopt an explicit-duration semi-Markov model for the state sequence $\{X_t\}$ [2]. That is, the patient's state evolves through a sequence of {\it super-states} $\{S_n\}, S_n \in \mathcal{X}, \forall n$, with each super-states lasting for a random duration $T_{n}$ which follows a distribution $T_{n}|S_n = s \sim f(T_{n} = \tau | \alpha_{s}),$ where $\alpha_{s}$ is the parameter of the duration distribution $f(.)$. We assume that $f(.)$ is a {\it negative binomial distribution}; the {\it geometric distribution} is a special case of the negative binomial distribution, and thus our model encapsulates Markovian transitions as a special case. The patient's state at time $t$ takes a value $X_t = S_n$ as long as the time step $t$ is occupied by the super-state duration $T_{n}$. The super-state transition probabilities are time-homogeneous and are given by $\mathbb{P}(S_{n+1} = j |S_{n} = i) = p_{ij}$. 

We assume that the super-states have a {\it gambler's ruin} structure as shown in Figure 1, i.e. $p_{ij} = 0, \forall |i-j|>1,$ and $p_{11}=p_{NN} = 1$. Such a structure is advantageous for the following reasons. First, it adds a semantic ingredient to the states that facilitate their interpretability, i.e. state $1$ is a clinical stability (absorbing) state, state $N$ is a clinical deterioration (absorbing) state, and all other states are transient states with intermediate, monotonically increasing levels of clinically severity. Second, such a structure facilitates unsupervised model learning when an informatively censored dataset is available, i.e. if physiological observations are recorded until one of the absorbing states materialize, and such state is declared in the dataset. The initial super-state probabilities are denoted as $\{p^{o}_{i}\}_{i=1}^{N},$ where $\sum_{i=1}^{N} p^{o}_{i} = 1$. Every patient's super-state sequence $\{S_n\}^{N_s}_{n=1}$ comprises $N_s$ super-states that are terminated by an absorbing states, i.e. $S_{N_s} \in \{1,N\}$, and hence $N_s$ the random number of clinical state transitions that a patient experiences during her hospitalization.

We call the physiological model described above a Semi-Markov Switching Linear Gaussian Model (SSLGM), and denote every instantiation of such a model as $\mathcal{M}(\Theta)$ with the parameter set $\Theta$ given by $\Theta = (\mathcal{X}, \{\Sigma_i, A_i, B_i, C_i, D_i, \alpha_i\}_{i=1}^{N}, \{p^{o}_{i}\}_{i=1}^{N}, \{p_{ij}\}_{i,j \in \mathcal{X}}).$

In order to capture the heterogeneity of the patients' population, we assume that the patients' physiological data are drawn from a mixture of SSLGM models with $G$ mixture components, i.e. $\{Y_t\}|Q = q \sim \sum_{g=1}^{G} w_g(q)\mathcal{M}(\Theta_{g}),$ where $w_g$ and $\Theta_g = (\mathcal{X}, \{\Sigma^g_i, A^g_i, B^g_i, C^g_i, D^g_i, \alpha^g_i\}_{i=1}^{N}, \{p^{o,g}_{i}\}_{i=1}^{N}, \{p^g_{ij}\}_{i,j \in \mathcal{X}})$ are the mixture weight and the parameter set for model $g$ respectively, whereas $Q$ is the patient's baseline admission information, i.e. the static information gathered about the patient upon admission to the ward, such as the age, gender, ICD9 code, etc. We assume that $w_g(q)$ is a linear function of $q$ for every $g$, and that the clinical state-space $\mathcal{X}$ is shared among all the models.  

\begin{figure}[t!]
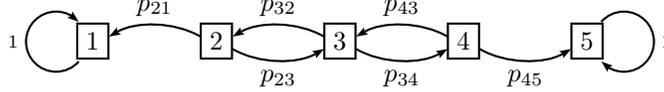

        \centering
        \psset{arrows=<-, arrowinset=0.15, shortput=nab, labelsep=2pt}
\[ \begin{psmatrix}[ colsep = 1.2]
    \psframebox{\pnode(0,3pt){S1} 1} & \psframebox{2} & \psframebox{3} & \psframebox{4} & \psframebox{5\pnode(0,3pt){S5}}
    \psset{offset=-1.5pt, nodesep =-1pt, arcangleA=30,arcangleB=30}
    \nccircle[angleA=90, nodesep=8pt]{S1}{0.4}\nbput{\scriptstyle1}
    \ncarc[nodesep = -1pt]{1,1}{1,2}^{p_{21}}\ncarc{1,2}{1,3}^{p_{32}} \ncarc{1,3}{1,4}^{p_{43}}
    \psset{offset=-0.5pt, nodesep=-0.5pt}
    \ncarc{1,3}{1,2}_{p_{23}}\ncarc{1,4}{1,3}_{p_{34}} \ncarc{1,5}{1,4}_{p_{45}}
    \nccircle[angleA=-90, nodesep=8pt]{S5}{0.4}\nbput{\scriptstyle1}
    \end{psmatrix} \]	
				\captionsetup{font= small}
        \caption{A clinical state model with 5 super-states.}
\label{Fig1}
\end{figure}	
 
\section{The Backward Labeling EM Algorithm}	  
In this section, we develop an algorithm for learning the mixture model $\{w_g(q), \mathcal{M}(\Theta_g)\}_{g=1}^{G}$ from an offline EHR dataset $\mathcal{D}$. Real-time inference of the clinical state for a newly hospitalized patient is then carried out using forward filtering in an online fashion. The offline dataset $\mathcal{D} = (\{Y_t\}^{J_k}_{t=1}, q_k, F_k)_{k=1}^{K}$ comprises the physiological data streams recorded for $K$ previously hospitalized patients, where $J_k$ is the amount of time the patient was hospitalized in the ward, $q_k$ is her admission information, and $F_k$ is the clinicians' intervention; $F_k = 0$ means that patient $k$ was discharged home, whereas $F_k = 1$ means that the patient was transferred to the ICU. We treat $F_k$ as a label for the absorbing state that has materialized for patient $k$ (i.e. $F_k = 0$ corresponds to state $1$ and $F_k = 1$ corresponds to state $N$). Algorithm 1 encapsulates the pseudocode for both the learning and inference algorithms. In these algorithms, we use the following divergence measure between two segments $X_1$ and $X_2$ of lengths $U_1$ and $U_2$ in a time series [3] 
\[\hat{\mathcal{E}}(X_1, X_2; \zeta) = \sum_{i=1}^{U_1} \sum_{j=1}^{U_2} \frac{2 ||X_{1,i}-X_{2,i}||^{\zeta}}{U_1 U_2} -\sum_{1\leq i \leq k \leq U_1}\frac{||X_{1,i}-X_{1,k}||^{\zeta}}{\binom{U_1}{2}}-\sum_{1\leq j \leq k \leq U_2}\frac{||X_{2,i}-X_{2,k}||^{\zeta}}{\binom{U_2}{2}},\]
 where $\zeta \in (0,2)$.
				
\begin{figure*}[t!]
\centering
  \centering
  \includegraphics[width=3.5in]{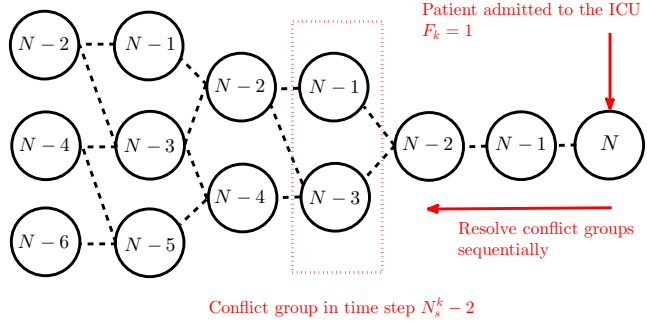}
				\captionsetup{font= small}
        \caption{Depiction for the operation of the backward labeling algorithm.}
\label{Fig2}
\end{figure*}

\begin{algorithm}[h]
  \caption{Offline Learning and Online Inference Algorithms}\label{alg1}
  \begin{algorithmic}[1]  
    \Procedure{\textsc{BackwardLabelingEM}}{$\mathcal{D} = (\{Y_t\}^{J_k}_{t=1}, q_k, F_k)_{k=1}^{K}$}
			  \State Detect the change points in $(\{Y_t\}^{J_k}_{t=1})_{k=1}^{K}$ using the non-parametric {\it E-divisive} and {\it E-Agglo} algorithms with the divergence measure $\hat{\mathcal{E}}$ to obtain estimates for the number of super-states $\hat{N}^k_s$ and their durations $(\{\hat{T}^k_n\}^{\hat{N}^k_s}_{n=1})^{K}_{k=1}$. 			
				\State Assign a label 1 to all the absorbing states ($\hat{S}^k_{\hat{N}^k_s} = 1$) for all patients with $F_k = 0$, and a label $N$ to all the absorbing states ($\hat{S}^k_{\hat{N}^k_s} = N$) for all patients with $F_k = 1$. 
				\State Apply backward state labeling for the transient states $(\{\hat{S}^k_{n}\}_{n=1}^{\hat{N}_{s}^{k}-1})_{k=1}^{K}$.
				\State Given the estimated semi-Markov process $\{(\hat{S}^k_{n}, \hat{T}^k_n)^{\hat{N}^k_s}_{n=1}\}^{K}_{k=1},$ estimate the model parameters $(\hat{\Theta}_g)_{g=1}^{G}$ using the standard EM algorithm.
    \EndProcedure
		\Procedure{\textsc{RiskScoring}}{$(Y_1,.\,.\,.,Y_t), \{\hat{w}_g(q), \mathcal{M}(\hat{\Theta}_g)\}_{g=1}^{G}$}
			  \State Estimate the latent factor $\hat{Z}_t$ using a {\it Kalman filter} via the estimated SSLGM parameters.
				\State Estimate the state sequence $\{\hat{X}_\tau\}_{\tau=1}^{t}$ using a {\it Rauch-Tung-Striebel} (RTS) smoother.
				\State Estimate the patient's risk score $R(t) = \mathbb{P}(X_{\infty} = N | Y_1,.\,.\,.,Y_t)$ using {\it forward filtering} for the estimated super-states and their durations up to time $t$.
    \EndProcedure
  \end{algorithmic}
\end{algorithm}	
	
The operation of the learning algorithms can be summarized as follows. Using the non-parametric E-divisive and E-Agglo algorithms, we can detect the change points in every patient's physiological sequence $\{Y^k_t\}$, and hence we can estimate the start and end times of every hidden super-state for those patients. Conditioned on the super-state durations, the super-state transitions reduce to an ordinary Markov process that is described by the Markov chain in Figure 1. Using the absorbing state label for every patient and exploiting the gambler's ruin transition structure, we can assign a label to the last three states in every patient's super-state sequence (e.g. the last three super-state are states $N-2$, $N-1$ and $N$ if $F_k=1$ as shown in Figure 2). The preceding states can then be labeled by creating {\it conflict groups} of potential states that could have materialized in a certain time slot, and selecting the one for which the divergence measure between the observations associated with that state with respect to a previously labeled state is minimized (e.g. in the conflict group highlighted in Figure 2, we resolve the conflict by picking state $N-1$ if the divergence measure with the observations associated with the previous labeled state $N-1$ at time slot $N_s-1$ is minimized). This {\it backward labeling} procedure proceeds until all states are labeled, and hence the SSLGM becomes a conditionally linear Gaussian model for which a direct application of the EM algorithm can allow learning the model parameters [4]. Online risk scoring is achieved by estimating the monitored patient's latent states $\{\hat{X}_\tau\}_{\tau=1}^{t}$ using the RTS smoothing algorithm, mapping those states to an estimated super-states sequence and durations, and then computing the probability of the super-states sequence being absorbed in state $N$.   

\section{Results}
We evaluated the utility of our model by conducting experiments on a heterogeneous cohort of 6,094 patients admitted to a general medicine floor in a large academic medical center in the period between March 2013 to February 2016. The cohort comprised patients with various ICD-9 codes corresponding to a wide variety of medical conditions (e.g. pneumonia, hematologic malignancies, sepsis, septicemia, etc). We trained a 3-state SSLGM model using a training set comprising the patients admitted in the period between March 2013 and July 2015 (4936 patients). The learned super-state transition parameters are $p^{o}_1 \approx 0.44, p^{o}_2 \approx 0.54, p^{o}_3 \approx 0.02, p_{21} \approx 0.94, p_{23} \approx 0.06$ and the estimated duration distribution for state 2 is
\[f(T_n = k| S_n = 2) = \frac{\Gamma(k+1.4541)}{k!\, \cdot\, \Gamma(1.4541)} \, \cdot \, (0.839)^{1.4541} \cdot \, (0.161)^{k}, k \in \mathbb{N}.\]   

\begin{table}[h]
\caption{\small Performance of various risk scoring methods ($^{\dagger}$LR: logistic regression. $^{\S}$RF: random forest).} 
\centering 
\begin{tabular}{c ccccccc}
\hline\hline 
{\bf \small Method} & {\bf \small SSLGM} & {\bf \small Rothman} & {\bf \small MEWS} & {\bf \small SOFA} & {\bf \small APACHE} & {\bf  \small LR$^{\dagger}$} & {\bf \small RF$^{\S}$}\\
\hline 
\centering
{\small \bf AUC (ICU)}  & {\small 0.47} & {\small 0.25} & {\small 0.18} & {\small 0.13} & {\small 0.13} & {\small 0.27}& {\small 0.36}\\ \hline  
{\small \bf AUC (ICU/discharge)}  & {\small 0.36} & {\small 0.25} & {\small 0.18} & {\small 0.1} & {\small 0.13} & {\small 0.17} & {\small 0.19}\\ \hline 
\end{tabular}
\label{tab:hresult}
\end{table}

\begin{figure}[t!]
\centering
  \centering
  \includegraphics[width=4.5in]{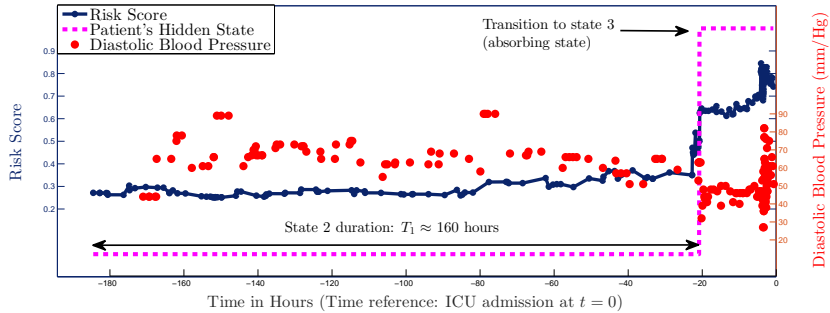}
				\captionsetup{font= small}
        \caption{Depiction for the operation of the inference and risk scoring algorithm.}
\label{Fig3}
\end{figure}
Table 1 demonstrates the AUC performance (TPR vs PPV) of the risk scoring methods based on the SSLGM model versus those based on state-of-the-art critical care risk scoring systems (Rothman index, MEWS, SOFA and APACHE), in addition to other benchmark algorithms (logistic regression and random forest). All algorithms were tested on the most recent patient records (1155 patients admitted between July 2015 and February 2016). The SSLGM model outperforms the Rothman index in predicting ICU admissions by $22\%$, and outperforms random forest by 11$\%$ ($p$-values $<$ 0.01). Similarly, the SSLGM model outperforms all other methods in predicting both ICU admissions and discharges from the ward. Figure 3 depicts the operation of the inference algorithm for one clinically deteriorating patient who ended up going to the ICU. As we can see, the proposed model predicts clinical deterioration for this representative patient 20 hours prior to the patient decompensating and requiring an emergent transfer to the ICU. The extra time afforded by this substantially earlier warning might allow the clinician sufficient time to prevent the patient from clinical deterioration The average timeliness of the predictions issued by the SSLGM model is 8 hours for a TPR of $50\%$ and a PPV of $35\%$.
\section*{References}
\small

\smallskip \noindent [1] Fox, Emily and Sudderth, Erik B. and Jordan, Michael I. and Willsky, Alan S. 2011. Bayesian nonparametric inference of switching dynamic linear models. \textit{IEEE Transactions on Signal Processing}, 59(4): 1569--1585. \\
\smallskip \noindent [2] Johnson, M. and Willsky, A. 2013. Bayesian nonparametric hidden semi-Markov models. \textit{JMLR}: 673--701.\\
\smallskip \noindent [3] Matteson, David S., and James, Nicholas A. 2014. A nonparametric approach for multiple change point analysis of multivariate data. \textit{Journal of the American Statistical Association}, 109(505): 334--345.\\
\smallskip \noindent [4] Wang, Xiang and Sontag, David and Wang, Fei. 2014. Unsupervised learning of disease progression models. \textit{ACM SIGKDD}: 85--94.
\end{document}